\documentclass[journal]{IEEEtran}

\usepackage{amsmath, cite, graphicx, multirow, amsfonts, booktabs, microtype, color}
\usepackage{amssymb}
\usepackage[colorlinks=false, hidelinks]{hyperref}
\usepackage[caption=false]{subfig}

\usepackage{pifont}
\usepackage{makecell}
\usepackage[switch]{lineno}
\switchlinenumbers 
\begin{document}
%
\title{Sparse Pedestrian Character Learning for Trajectory Prediction}
%
%
%

\author{Yonghao Dong,
        Le~Wang,~\IEEEmembership{Senior Member,~IEEE,}
        Sanpin~Zhou,~\IEEEmembership{Member,~IEEE,}
        Gang~Hua,~\IEEEmembership{Fellow,~IEEE,}
       	and~Changyin Sun,~\IEEEmembership{Senior Member,~IEEE}
\thanks{Manuscript received xxx; revised xxx; accepted xxx. Date of publication xxx; date of current version xxx. This work was supported in part by National Key R\&D Program of China under Grant 2021YFB1714700, NSFC under Grants 62088102 and 62106192, Natural Science Foundation of Shaanxi Province under Grants 2022JC-41 and 2021JQ-054, China Postdoctoral Science Foundation under Grant 2020M683490, and Fundamental Research Funds for the Central Universities under Grants XTR042021005 and XTR072022001. \textit{(Corresponding author: Le Wang.)}}%
\thanks{Yonghao Dong, Le Wang and Sanping Zhou are with the National Key Laboratory of Human-Machine Hybrid Augmented Intelligence, National Engineering Research Center for Visual Information and Applications, and Institute of Artificial Intelligence and Robotics, Xi'an Jiaotong University, Xi'an, Shaanxi 710049, China. (e-mail: yhdong@stu.xjtu.edu.cn; \{lewang, spzhou\}@mail.xjtu.edu.cn)}
\thanks{Gang Hua is with Wormpex AI Research, Bellevue, WA 98004, USA. (e-mail: ganghua@gmail.com)}
\thanks{Chanyin Sun is with School of Artificial Intelligence, Anhui University, Hefei, Anhui 230039, China. (e-mail: cysun@ahu.edu.cn)}}

\maketitle

\begin{abstract}
Pedestrian trajectory prediction in a first-person view has recently attracted much attention due to its importance in autonomous driving. 
Recent work utilizes pedestrian character information, \textit{i.e.}, action and appearance, to improve the learned trajectory embedding and achieves state-of-the-art performance.
However, it neglects the invalid and negative pedestrian character information, which is harmful to trajectory representation and thus leads to performance degradation.
To address this issue, we present a two-stream sparse-character-based network~(TSNet) for pedestrian trajectory prediction. 
Specifically, TSNet learns the negative-removed characters in the sparse character representation stream to improve the trajectory embedding obtained in the trajectory representation stream.
Moreover, to model the negative-removed characters, we propose a novel sparse character graph, including the sparse category and sparse temporal character graphs, to learn the different effects of various characters in category and temporal dimensions, respectively.
Extensive experiments on two first-person view datasets, PIE and JAAD, show that our method outperforms existing state-of-the-art methods. In addition, ablation studies demonstrate different effects of various characters and prove that TSNet outperforms approaches without eliminating negative characters.
\end{abstract}


%
\IEEEpeerreviewmaketitle

\section{Introduction}
\label{sec:introduction}

\begin{figure}[t]
	\centering
	\includegraphics[width=\linewidth]{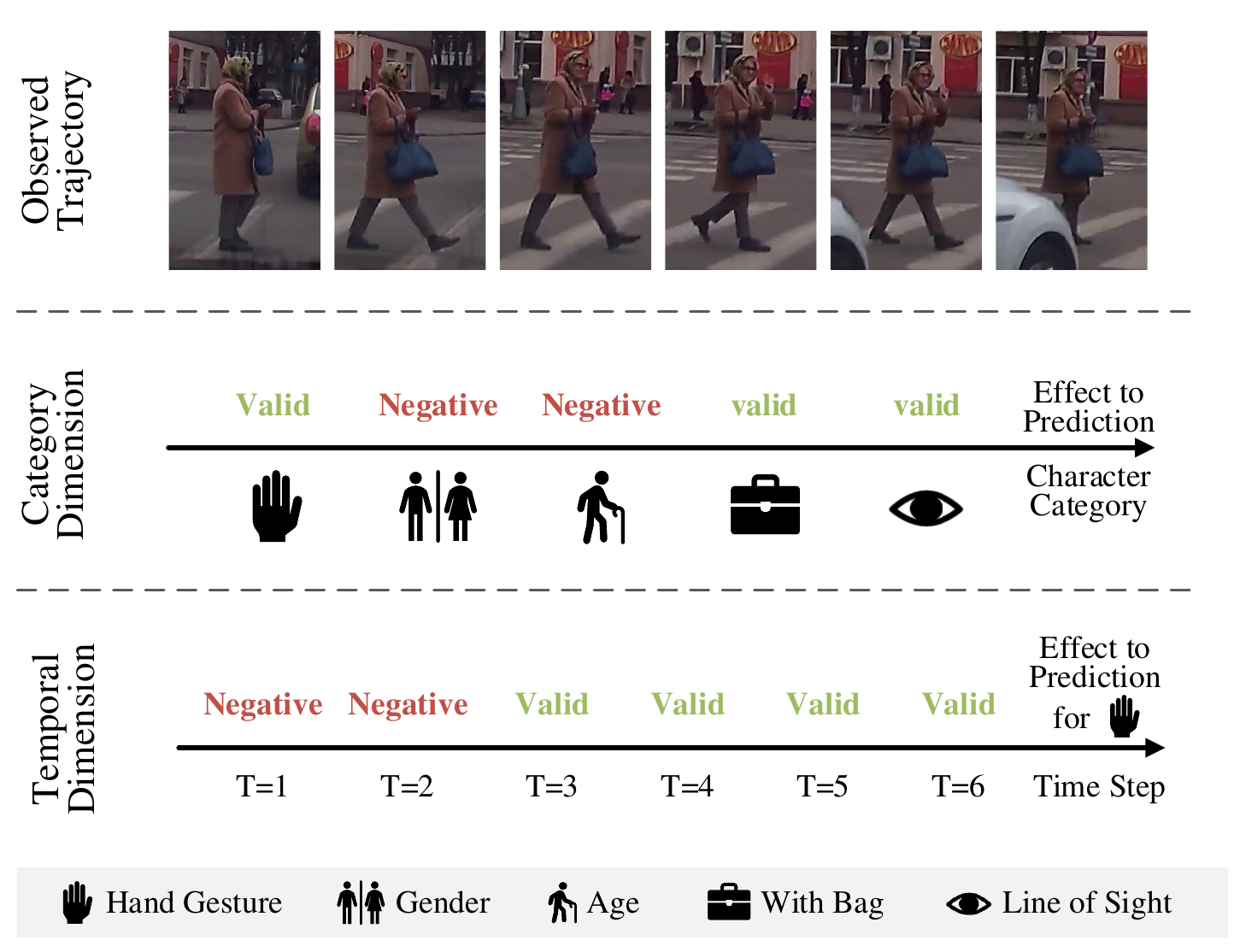}
	\caption{
		Illustration of negative characters in category and temporal dimensions. 
		In the category dimension, some pedestrian characters are irrelevant to trajectory modes~(\textit{i.e.}, gender, age). 
		In the temporal dimension, a determined type of character~(such as the hand gesture) at time steps far from future~(\textit{i.e.}, $T=1, 2$) are irrelevant to trajectory modes.
		We name the character information irrelevant to trajectory modes as negative characters, which may negatively affect trajectory representation and lead to performance degradation.
	}
	\label{fig1}
\end{figure}

\IEEEPARstart{P}{edestrian} trajectory prediction in a first-person view is a crucial technology in autonomous driving~\cite{sun2022m2i,ren2021safety,wen2022social,gu2021densetnt,zhang2022d2,wang2021multi,hu2023holistic} because it is necessary to predict nearby pedestrians’ actions and locations to avoid collision. 
Unlike pedestrian trajectory prediction in a bird-eye view camera~(BEV)~\cite{tsao2022social, monti2022many, wong2022view, dendorfer2021mg, chen2021personalized}, which mainly focuses on trajectory coordinates, pedestrian trajectory prediction in a first-person view camera~(FPV)~\cite{halawa2022action,yao2021bitrap,neumann2021pedestrian,yao2019egocentric} have richer annotated pedestrian character information at each time step, including their actions, gestures, genders, etc. They can help distinguish some similar trajectories that are indistinguishable by trajectory coordinates.
For example, with the same historical trajectory and final goal, the future trajectory of a girl texting on the phone while walking could be different from a man pushing a stroller~\cite{halawa2022action}.

Previous work~\cite{halawa2022action} has proved that pedestrian character information, \textit{i.e.}, action, can improve the accuracy of trajectory prediction by improving the trajectory coordinates representation.
However, it ignores invalid or negative pedestrian character information, which could cause performance degradation, especially regarding multi-category characters and long-term prediction.
As shown in Figure~\ref{fig1}, in the category dimension, different kinds of pedestrian characters, such as hand gestures, gender, age, etc., influence performance differently due to whether they are relevant to future trajectory modes.
Some types of pedestrian characters have positive effects on performance for trajectory prediction, while some may have adverse effects, for example, the prediction using ``gender" or ``age" character information in some datasets, which is proved by our ablation study. 
In the temporal dimension, pedestrian character information at different time steps has different influences on performance because of different time intervals to the future.
Pedestrian characters at some distant time steps from the future may be useless or even have adverse effects, for example, time steps at $T=1, 2$.
In sum, some pedestrian characters can improve prediction accuracy to varying degrees, while some may cause negative influences.
Hence, it is a meaningful task to make full use of valid pedestrian characters and eliminate negative ones.

\begin{figure*}[t]
	\centering
	\includegraphics[width=1.0\textwidth]{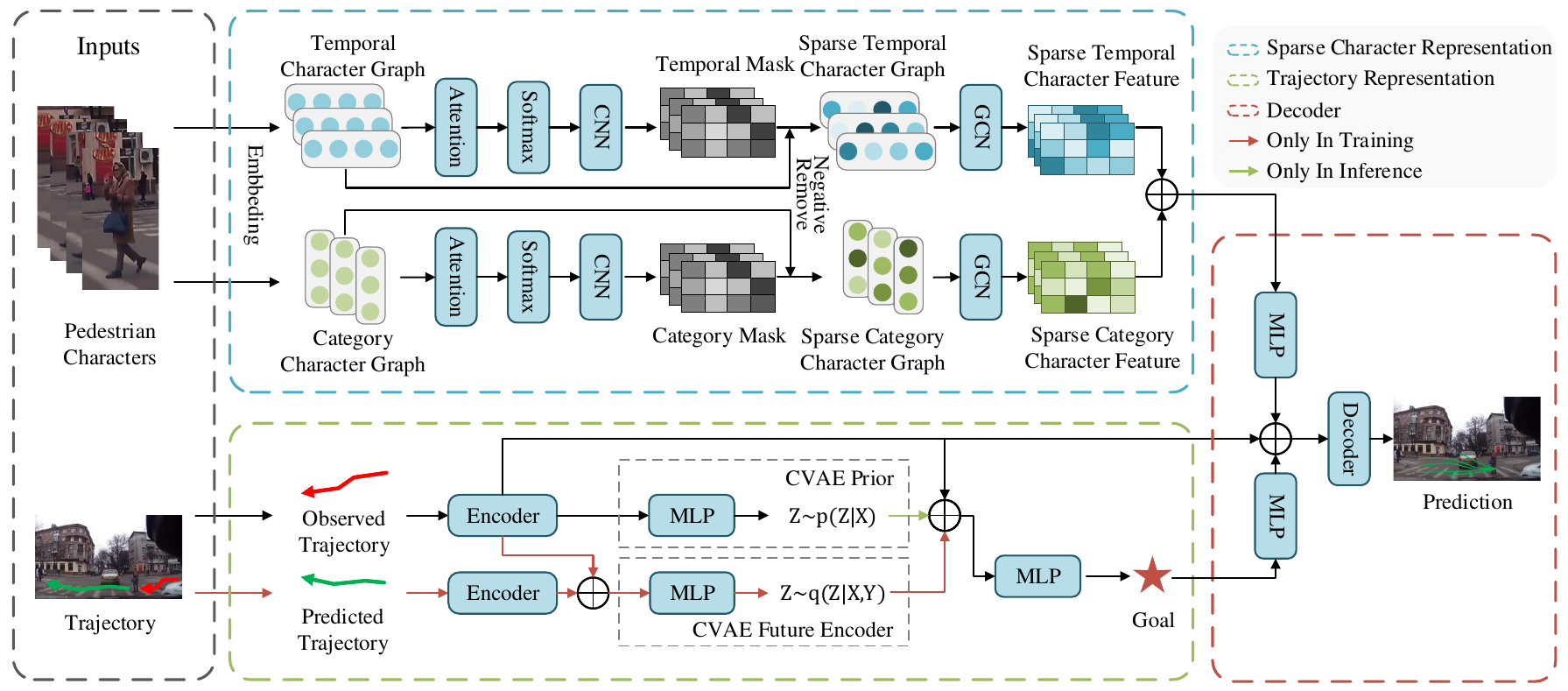}	
	\caption{  
		An overview of our proposed method.
		Our method consists of three key components: 
		(a) A sparse character representation stream, which learns the sparse character features from the pedestrian character inputs. 
		(b) A trajectory representation stream, which encodes the trajectory coordinates features. 
		(c) A decoder module, which decodes the concatenation feature from (a) and (b).
	}
	\label{fig2}
\end{figure*}

Motivated by the analysis above, we present a \textit{two-stream sparse-character-based network} for pedestrian trajectory prediction.
The proposed TSNet includes three key components: a sparse character representation stream, a trajectory representation stream, and a decoder module, as shown in Figure~\ref{fig2}. 
To model the negative-removed pedestrian characters, we propose \textit{a novel sparse character graph} to represent different effects of various pedestrian characters and remove side-effect ones in the sparse character representation stream.
The sparse character graph includes a sparse temporal character graph and a sparse category character graph to model different effects in temporal and category dimensions, respectively.
To construct the sparse temporal character graph without negative characters, we first learn the importance weights of a single character category at different time steps to form a mask. Then, by stacking the obtained masks of all categories together, we have a temporal mask of all characters. Finally, we use all characters and the temporal mask to generate a sparse temporal character graph without negative characters.
Similarly, to construct the sparse category character graph without negative characters, we can first learn the importance weights of all character categories at a single time step to form a mask, and then obtain the category mask of all characters by stacking masks of all time steps together. Then we can represent the negative-removed characters in the category dimension by a sparse category character graph formed by all characters and the obtained category mask.

By our proposed sparse character graph, we can learn the negative-removed character features in the sparse character representation stream using the self-attention mechanism and convolutional networks.
Meanwhile, we can learn the trajectory coordinates representation from observed trajectory in the trajectory representation stream by gated-recurrent unit (GRU) encoders.
Subsequently, we use the learned negative-removed character features to improve the trajectory representation by concatenation. 
Finally, we decode the improved trajectory representation into the predicted trajectory in the decoder module.
In summary, our contributions are four-fold: 
\begin{itemize}
	\item To the best of our knowledge, this is the first work that models sparse pedestrian characters in pedestrian trajectory prediction to make full use of valid characters and eliminate negative ones.
	\item We design a two-stream sparse-character-based pedestrian trajectory prediction network to improve the trajectory representation by negative-removed characters.
	\item We propose a novel sparse character graph for trajectory prediction to model the negative-removed representation of pedestrian characters according to its relevance to trajectory modes.	
	\item Experiments on well-established first-person view datasets demonstrate that our approach significantly outperforms the state-of-the-art methods. We also conduct extensive ablation studies to validate the effectiveness of our contributions.
\end{itemize}

The rest of this paper is organized as follows. 
We briefly review the related work in Section~\ref{Related Works}. 
We present the technical details of our proposed method in Section~\ref{Our Method}. 
Then extensive experiments and analysis are presented in Section~\ref{Experimental Analysis}. 
Finally, we conclude the paper in Section~\ref{Conclusion}.

\section{Related Works}
\label{Related Works}
In this section, we review previous works related to ours, which we categorize into three parts: (1) pedestrian trajectory prediction, (2) trajectory prediction using pedestrian characters and (3) graph structure learning.
\subsection{Pedestrian Trajectory Prediction }
Pedestrian trajectory prediction in a bird-eye view (BEV)~\cite{mohamed2022social,shafiee2021introvert,zamboni2022pedestrian,xia2022cscnet,pei2019human,wang2022spatio} observes the pedestrian trajectories from a static camera facing downwards. 
Earlier methods~\cite{alahi2016social,gupta2018social,shi2022social} only utilize trajectory coordinates as temporal sequences to make the prediction. Social-LSTM~\cite{alahi2016social} forecasts the pedestrian trajectory by learning coordinates representation with Recurrent Neural Networks. Social-GAN~\cite{gupta2018social} predicts the pedestrian trajectory by learning coordinates representation with the Generative Adversarial Network. SIT~\cite{shi2022social} models the human trajectory coordinates as a social interpretable tree to model the multimodal futures. 
However, people are not independent individuals in social environments, and the social interaction between people will also affect the future trajectories.
Hence, it is essential to consider human-human interaction~\cite{kosaraju2019social,yuan2021agentformer,bae2022learning} in pedestrian trajectory prediction. Some methods model interactions between humans to improve the future forecasting. Social-BiGAT~\cite{kosaraju2019social} models the human interaction by the graph attention network. AgentFormer~\cite{yuan2021agentformer} models the spatial interaction by the Transformer. GP-Graph~\cite{bae2022learning} models the interaction of pedestrians as a group graph. 
Since trajectory information is limited, many methods utilize scene information to help the prediction. Sophie~\cite{sadeghian2019sophie} models the scene with CNN~\cite{wang2021smsnet,wang2022deep}. Y-net~\cite{mangalam2021goals} models the scene using a series of U-nets~\cite{ronneberger2015u}. End-to-End~\cite{guo2022end} models the scene using CNN and a convolutional long short-term memory~\cite{shi2015convolutional} network. 

However, limited by the scene-invariant~\cite{rasouli2021bifold,yue2022human,bae2022non, cao2022advdo, pang2021trajectory} property of the bird-eye view, the practical scene information that can be used is minimal. What is worse, due to limitations of the bird-eye view, the BEV camera can not capture pedestrians' character information, such as human appearances and individual actions, which are more valuable for learning. 
Hence, more recent studies have begun to focus on the first-person view pedestrian trajectory prediction.
In this work, we focus on improving the trajectory coordinates representation through the proposed sparse character graph in the first-person view camera.

\subsection{Trajectory Prediction Using Pedestrian Characters}
Pedestrian Trajectory prediction in a first-person view (FPV)~\cite{halawa2022action,yao2021bitrap,neumann2021pedestrian,yao2019egocentric} observes the pedestrian trajectories from an on-board camera with ego-motion, which can capture more valuable pedestrian character information such as hand gesture, head motion, walking speed, age, gender, etc. 
Some works utilize individual visual features to improve pedestrian trajectory prediction in the first-person camera.
DBN~\cite{kooij2014context} models pedestrians' awareness from their faces by a Dynamic Bayesian Network to predict whether they cross the road.
FPL~\cite{yagi2018future} models pedestrian' body keypoint features by a convolutional neural network to improve the trajectory coordinate representation.
A recent work, ABC~\cite{halawa2022action}, proposes an action-based contrastive learning method to utilize pedestrian action information to improve trajectory representation by classifying similar trajectories with different hunman actions, which achieves state-of-the-art performance.
All the approaches demonstrate the effectiveness of pedestrian trajectory prediction with human characters.

However, previous works ignore the invalid or negative pedestrian characters, which could influence trajectory representation and thus cause performance degradation.
To address this issue, our proposed pedestrian sparse character graph can distinguish different effects of pedestrian characters in temporal and category dimensions. Accordingly, we can fully use valid pedestrian characters and eliminate invalid or negative ones.

\subsection{Graph Structure Learning}
Graph Structure Learning~(GSL)~\cite{xia2021self,zhao2021intrinsic,cao2022progressive,jian2013discriminative,mesgaran2020anisotropic} aim to learn a graph structure which is more suitable for a certain task from the original data feature and explore the implicit and correlative information absent from the original data~\cite{JIANG2023109265}.
The existing graph structure learning methods can be grouped into two categories: traditional graph methods and deep graph methods. 
For traditional graph methods, Takai et al.~\cite{takai2020hypergraph} developed local and global clustering algorithms based on PageRank for hypergraphs.
For deep graph methods, Social-STGCNN~\cite{mohamed2020social} propose a social spatio-temporal graph convolutional neural network to model the social interaction of pedestrians.
RSBG~\cite{sun2020recursive} recursively extract social representations supervised by group-based annotations and formulate them into a social behavior graph.
GroupNet~\cite{xu2022groupnet} propose a trainable multiscale hypergraph to capture both pair-wise and group-wise interactions at multiple group sizes.

The previous graph structure learning method aims to discover the implicit information. 
However, to fully use valid pedestrian characters and eliminate negative ones, this work aims to remove negative information which has side effects on the pedestrian trajectory representation.
Hence, we propose the sparse character graph to remove such negative information by eliminating nodes with the adaptive masks.
And the constructed sparse character graph can improve trajectory embedding without side-effect from negative characters.

\section{Method}
\label{Our Method}
In this section, we present our method for the first-person view pedestrian trajectory prediction, in which we focus on learning the negative-removed characters of pedestrians to improve the trajectory representation. We achieved this by constructing a two-stream sparse-character-based network, in which we model the past trajectory coordinates by the trajectory representation stream and then improve its learned representation by our proposed sparse character graph in the sparse character representation stream.

Section III-A first gives the definition of the pedestrian trajectory prediction problem. 
Section III-B details the sparse character representation stream, including the construction and generation of temporal \& category character graphs, temporal \& category masks and sparse temporal \& category graphs.
Section III-C details the trajectory representation stream, including trajectory encoders and the CVAE sampler.
Section III-D details the decoder module loss function of the overall framework.

\subsection{Problem Formulation}
Given an observed trajectory $X = [x_1, x_2, ..., x_{T_{obs}}]$, pedestrian trajectory prediction aims to predict a future trajectory $Y = [y_{T_{obs}+1}, y_{T_{obs}+2}, ..., y_{T_{pred}}]$, where $x$ and $y$ are bounding box coordinates of the observed trajectory and future trajectory, respectively. $T_{obs}$ and $T_{pred}$ are the maximum time lengths of the observed trajectory and predicted trajectory, respectively. In addition, we also have pedestrian character labels of $N$ categories for each observed trajectory $X$: $S = [s^1, s^2, ...,s^N]$ at each time step, where $s^n = [s_1^n, s_2^n, ..., s_{T_{obs}}^n]$. Note that $s_t^n$ is the pedestrian character information of the $n^{th}$ category at the $t^{th}$ observed time step.

\subsection{Sparse Character Representation Stream}
\textbf{Temporal \& Category Character Graphs. }
To model the pedestrian characters, we first construct the temporal and category character graphs. 
Given the pedestrian character input $S$, we can construct a set of temporal character graphs $G_{n}^{tem} \in \{ G_{1}^{tem}, G_{2}^{tem}, ..., G_{N}^{tem} \}$, which represent the pedestrian character information of category $n$ at all observed time steps.
We defined the graph $G_{n}^{tem} = (V_n, E_n, F_n)$, where $V_n = \{v_i^n | i = \{ 1,...,T_{obs} \} \}$ is the vertex set of the graph $G_{n}^{tem}$. The pedestrian character information $s_i^n$ is the attribute of $v_i^n$.
$E_n = \{e_{i,j}^n |  i,j = \{ 1,...,T_{obs} \} \}$ is the edge set of the graph $G_{n}^{tem}$. 
$F_n = \{ f_{i}^n|  i = \{ 1,...,T_{obs} \} \} \in \mathbb{R}^{T_{obs} \times D_f^n} $ is the feature matrix associated with each pedestrian character information $v_i^n$, where $ D_f^n$ is the feature dimension.

Similarly, we can construct a set of category character graphs of pedestrians $G_{t}^{cat} \in \{ G_{1}^{cat}, G_{2}^{cat}, ..., G_{T_{obs}}^{cat} \}$, which represent all categories' pedestrian character information at time step $t$. It is define as $G_{t}^{cat} = (V_t, E_t, F_t)$, where $V_t = \{v_t^i |  i = \{ 1,...,N \} \}$ is the vertex set of the graph $G_{t}^{cat}$. The pedestrian character information $s_t^i$ is the attribute of $v_t^i$.
$E_t = \{e_t^{i,j} | i,j = \{ 1,...,N \} \}$ is the edge set of the graph $G_{n}^{cat}$. 
$F_t = \{f_t^{i}| i = \{ 1,...,N \} \} \in \mathbb{R}^{N \times D_f^t}$ is feature matrix associated with each pedestrian character information $v_t^i$, where $ D_f^t$ is the feature dimension. 
We initialize all attributes $e_{i,j}^n$ and $e_t^{i,j}$ as one, assuming that each pair of vertices is connected by an edge.

The topological structures of graphs $G_{n}^{tem}$ and $G_{t}^{cat}$ are represented by the adjacency matrices $A_n = \{ a_{n;i,j} | i,j=1,...T_{obs} \} \in \mathbb{R}^{T_{obs} \times T_{obs}}$ and $A_t = \{ a_{t;i,j} | i,j=1,...N \} \in \mathbb{R}^{N \times N}$, respectively.
The values of $a_{n;i,j}$ and $a_{t;i,j}$ in adjacency matrices $A_n$ and $A_t$ are initialized as follows:
\begin{equation}
	\begin{split} \label{eq0}
		a_{n;i,j} &= ||v_i^n - v_j^n||_2,\\
		a_{t;i,j} &= ||v_t^i - v_t^j||_2,
	\end{split}
\end{equation} 
where $||*||_2$ is $L_2$-norm.
The values of $f_{i}^n$ and $f_t^{i}$ in feature matrix $F_n$ and $F_t$ are defined as:
\begin{equation}
	\begin{split} \label{eq1}
		f_{i}^n &= \phi (v_i^n, W_n),\\
		f_t^{i} &= \phi (v_t^i, W_t), 
	\end{split}
\end{equation} 
where $\phi(\cdot,\cdot)$ denotes linear transformation. $ W_n$ and $W_t$ are the weights of the linear transformation.

\textbf{Temporal \& Category Masks. }
To distinguish the different effects of pedestrian character in temporal and category dimensions, we need to learn the sparsity masks of the pedestrian character features $F_n$ and $F_t$. 
We first compute the temporal character attention score matrix $\mathbf{R}_t \in \mathbb{R}^{H \times T_{obs} \times T_{obs}}$ by the multi-head self-attention mechanism~\cite{fu2019dual,zhang2022resnest} as:
\begin{equation}
	\begin{split} \label{eq2}
		\mathbf{Q}_i^{tem} &= \phi (F_n, W_Q^{t}),\\
		\mathbf{K}_i^{tem} &= \phi (F_n, W_K^{t}), \\
		\mathbf{O}_{i}^{tem} &= \text{Softmax}(\frac{\mathbf{Q}_i^{tem} (\mathbf{K}_i^{tem})^T}{\sqrt{d_t}}), \\
		\mathbf{R}_{t} &= \text{Concat}(\mathbf{O}_{i}^{tem}), i=1,2,...,H,
	\end{split}
\end{equation} 
where $\mathbf{Q}_i^{tem} \in \mathbb{R}^{T_{obs} \times D_q^t}$ and $\mathbf{K}_i^{tem} \in \mathbb{R}^{T_{obs} \times D_k^t}$ are the query and key, respectively. $W_Q^{t}$ and $W_K^{t}$  are the weights of the linear transformation. $i$ is the index of $H$ heads. $\mathbf{O}_{i}^{tem} \in \mathbb{R}^{T_{obs} \times T_{obs}}$ is the attention score of the $i^{th}$ attention head. $\sqrt{d_t} = \sqrt{D_q^t}$ is a scaled factor~\cite{vaswani2017attention}.
Similarly, we can compute category character attention score matrix $\mathbf{R}_c \in \mathbb{R}^{H \times N \times N}$ as:
\begin{equation}
	\begin{split} \label{eq3}
		\mathbf{Q}_i^{cat} &= \phi (F_{t}, W_Q^{c}),\\
		\mathbf{K}_i^{cat} &= \phi (F_{t}, W_K^{c}), \\
		\mathbf{O}_{i}^{cat} &= \text{Softmax}(\frac{\mathbf{Q}_i^{cat}(\mathbf{K}_i^{cat})^T}{\sqrt{d_c}}),\\
		\mathbf{R}_{c} &= \text{Concat}(\mathbf{O}_{i}^{cat}), i=1,2,...,H,
	\end{split}
\end{equation} 
where $\mathbf{Q}_i^{cat} \in \mathbb{R}^{N \times D_q^c}$ and $\mathbf{K}_i^{cat} \in \mathbb{R}^{N \times D_k^c}$ are the query and key, respectively. $W_Q^{c}$ and $W_K^{c}$  are the weights of the linear transformation. $i$ is the index of $H$ heads. $\mathbf{O}_{i}^{cat} \in \mathbb{R}^{T_{obs} \times T_{obs}}$ is the attention score of the $i^{th}$ attention head. $\sqrt{d_c} = \sqrt{D_q^c}$ is a scaled factor~\cite{vaswani2017attention}.

Since the multi-head attention scores come from different representation subspaces, we fuse it through a convolution network~\cite{choy20194d} and then adopt a sigmoid function to map the attention scores to mask values $[0,1]$, as follows:
\begin{equation}
	\begin{split} \label{eq4}
		\mathbf{J}_t &= \delta \ (\text{Conv} \ (\mathbf{R}_t,\mathcal{K})), \\
		\mathbf{J}_c &= \delta \ (\text{Conv} \ (\mathbf{R}_c,\mathcal{K})),
	\end{split}
\end{equation}
where $\mathbf{J}_t \in \mathbb{R}^{T_{obs} \times T_{obs}}$ and $\mathbf{J}_c \in \mathbb{R}^{N \times N}$ are the feature maps of $\mathbf{R}_t$ and $\mathbf{R}_c$, respectively. $\delta(\cdot)$ is a sigmoid function. $\mathcal{K}$ denotes the $1 \times 1$ convolution kernel.

To model the different pedestrian characters' effects and discard invalid or negative ones in pedestrian character graphs, we generate sparse masks $M_n$ and $M_t$ of pedestrian character graphs $G_n^{tem}$ and $G_t^{cat}$ by an element-wise threshold $\xi$.
When the element in $\mathbf{J}_t$ or $\mathbf{J}_c$ is larger than $\xi$, we do not change its value, otherwise we set it to zero, as follows:
\begin{equation}
	\begin{split} \label{eq5}
		M_n &= \mathcal{I}(\mathbf{J}_t \ge \xi), \\
		M_t &= \mathcal{I}(\mathbf{J}_c \ge \xi),
	\end{split}
\end{equation}
where $\mathcal{I}(a \ge b)$ is an indicator function, which denotes that elements in $a$ keep their values if the inequality holds, otherwise change their values to zero.

\textbf{Sparse Temporal \& Category Character Graphs. }
Since we have obtained the sparse masks $M_n$ and $M_t$ which represent different pedestrian characters' effects, we can generate the sparse character features as follows:
\begin{equation}
	\begin{split} \label{eq6}
		\hat{F}_n &= \text{Softmax}(F_n \odot M_n), \\
		\hat{F}_t &= \text{Softmax}(F_t \odot M_t),
	\end{split}
\end{equation}
where $\hat{F}_n$ and $\hat{F}_t$ are the sparse temporal and sparse category character feature matrices, respectively. $\odot$ denotes the element-wise multiplication. 
Hence, we can obtained the sparse temporal character graph $\hat{G}_n^{tem} = (V_n,E_n,\hat{F}_n)$ and the sparse category character graph $\hat{G}_t^{cat} = (V_t,E_t,\hat{F}_t)$.

Then, we adopt a graph convolutional network (GCN)~\cite{chen2020simple, wu2019simplifying} to obtain the high-level features of the sparse graphs. We first add identity matrices to the adjacency matrices $A_n$ and $A_t$ following previous methods~\cite{mohamed2020social, xu2022adaptive}, as follows:
\begin{equation}
	\begin{split} \label{eq7}
		A'_n &= A_n + I,  \\
		A'_t &= A_t + I.
	\end{split}
\end{equation}
Secondly, we stack $A'_n$ from all pedestrian character categories as $\hat{A}_n=\{ A'_1, A'_2,...,A'_N \} \in \mathbb{R}^{N \times T_{obs} \times T_{obs}}$ and stack $A'_t$ from all time steps as $\hat{A}_t=\{ A'_1, A'_2,...,A'_{T_{obs}} \} \in \mathbb{R}^{T_{obs} \times N \times N}$.
Then, we stack feature matrices of the $l^{th}$ layer as $F_n^{(l)}=\{ F_1^{(l)},F_2^{(l)},...,F_N^{(l)} \} \in \mathbb{R}^{N \times D_f^n  \times T_{obs}}$ and $F_t^{(l)}=\{ F_1^{(l)},F_2^{(l)},...,F_{T{obs}}^{(l)} \} \in \mathbb{R}^{T_{obs} \times D_f^t  \times N}$, respectively.
We also stack node degree matrices $D_n=\{ D_1,D_2,...,D_N \}$ and $D_t=\{ D_1,D_2,...,D_{T_{obs}} \}$, respectively.
Finally, we have output features $\mathbf{F}_n^{(l+1)}\in \mathbb{R}^{N \times D_f^n  \times T_{obs}}$ and $\mathbf{F}_t^{(l+1)} \in \mathbb{R}^{T_{obs} \times D_f^t  \times N}$ of the $(l+1)^{th} $ layer of the GCN, as follows:
\begin{equation}
	\begin{split} \label{eq8}
		\mathbf{F}_n^{(l+1)} &= \sigma (D_n^{-\frac{1}{2}} \hat{A}_n D_n^{\frac{1}{2}} \mathbf{F}_n^{(l)} W_n^{(l)}), \\
		\mathbf{F}_t^{(l+1)} &= \sigma (D_t^{-\frac{1}{2}} \hat{A}_t D_t^{\frac{1}{2}} \mathbf{F}_t^{(l)} W_t^{(l)}),
	\end{split}
\end{equation}
where $\sigma(\cdot)$ is a non-linearity activation function. 

\subsection{Trajectory Representation Stream}
The trajectory representation stream is constructed to obtain observed trajectory coordinates representations, which can be improved by the learned sparse character representations.
To generate multimodal trajectories, we utilize the CVAE sampler to select multiple goals, which guide the multimodal pedestrian trajectory prediction.

\textbf{Trajectory Encoders. } Human trajectories coordinates are time series information. To obtain the high dimensional representation, we use gated-recurrent unit~(GRU)~\cite{dey2017gate} encoders to extract trajectory features.
The past trajectory $X$ is processed by a GRU encoder to obtain the past trajectory feature $\mathbf{F}_p$.
Meanwhile, the ground truth $Y$ is processed by a GRU encoder to obtain the future trajectory feature $\mathbf{F}_g$, which only exists in the training process, as follows:
 \begin{equation}
 	\begin{split} 
 		\mathbf{F}_p = g_{enc}(X), \\
 		\mathbf{F}_g = g_{enc}(Y), 
 	\end{split}
 \end{equation}
where $g_{enc}(\cdot)$ denotes the GRU encoder.

\textbf{CVAE Sampler. }
To generate multimodal trajectories, we use CVAE to generate multimodal pedestrian trajectories by sampling variables in the latent space.
We first introduce a latent variable $Z\sim\mathcal{N}(\mu_Z, \sigma_Z)$. 
The probability density function $p(Z|X)$ predicts Gaussian parameters $\mu_{Z_p}$ and $\sigma_{Z_p}$ by $\mathbf{F}_p$.
The probability density function $q(Z|X,Y)$ predicts Gaussian parameters $\mu_{Z_q}$ and $\sigma_{Z_q}$ by both $\mathbf{F}_p$ and $\mathbf{F}_g$, as follows:
 \begin{equation}
	\begin{split} 
		q(Z|X,Y) = \mathcal{N}(\mu_{Z_q}, \sigma_{Z_q}), \\
		p(Z|X) = \mathcal{N}(\mu_{Z_p}, \sigma_{Z_p}), \\
	\end{split}
\end{equation}
In the training process, we can sample multiple $Z$ from $q(Z|X,Y)$ to generate multiple goals, and then generate multimodal trajectories.
However, we do not have future trajectories $Y$ during inference.
Hence, we optimize $p(Z|X)$ to approach $q(Z|X,Y)$ by the Kullback–Leibler divergence~(KLD) loss, thus we can sample multiple $Z$ from $q(Z|X)$ to generate multimodal trajectories during inference.
The KLD loss is shown as follows:
 \begin{equation}
	\begin{split} 
		\mathcal{L}_{KLD}=\sum{p(z) \times [log(\frac{p(z)}{q(z)})]}, \\
	\end{split}
\end{equation}

Therefore, in the training process, we can use $\mathbf{F}_p$ and $\mathbf{F}_g$ to train the CVAE to generate latent variables $Z$, which generate multimodal goals jointly with $\mathbf{F}_p$.
In the inference process, we can only use $\mathbf{F}_p$ to generate latent variables $Z$, which are concatenated with $\mathbf{F}_p$ to generate multimodal goals.

\subsection{Decoder and Loss Function}
To obtain the multi-modal trajectory predictions, we follow the previous best-of-$K$ \cite{halawa2022action, yao2021bitrap, zhao2021you, sun2021three} strategy by sampling $K$ latent variables $Z$ by CVAE and generating corresponding goals $G$. 
Then, we use a GRU decoder to predict the multi-modal trajectories $\hat{Y}$ by the learned sparse pedestrian character features and the multi-modal goals, as follows:
\begin{equation}
	\begin{split} \label{eq9}
		\hat{Y} = g_{dec}(\epsilon(\mathbf{F}_n^{(l+1)} \oplus \mathbf{F}_t^{(l+1)}) \oplus \mathbf{F}_p \oplus \epsilon(G)), 
	\end{split}
\end{equation}
where $\epsilon$ denotes an MLP layer. $\oplus$ is the concatenation function and $g_{dec}(\cdot)$ denotes the GRU decoder.

Our model is trained end-to-end by minimizing the loss function $\mathcal{L}_{TSNet}$ as:
\begin{equation}
	\begin{split} \label{eq10}
		\mathcal{L}_{TRJ} &= \min_{k \in K}||\hat{Y}^{(k)} - Y||_2,\\
		\mathcal{L}_{GL} &=  \min_{k \in K}||\hat{G}^{(k)} - G_{gt}||_2,\\
		\mathcal{L}_{TSNet} &= \mathcal{L}_{TRJ} + \mathcal{L}_{GL} + \mathcal{L}_{KLD}, 
	\end{split}
\end{equation}
where $\mathcal{L}_{TRJ}$ means the trajectory $L_2$-norm loss for the complete training process. $\mathcal{L}_{GL}$ means the $L_2$-norm loss between generated goals and the ground truth. $\mathcal{L}_{KLD}$ means the Kullback–Leibler divergence~(KLD) loss for CVAE. $G_{gt}$ is the ground truth of goals.

\begin{table*}[t]
	\caption{The quantitative results on two public benchmark datasets PIE and JAAD. All approaches input 15 frames and output 45 frames. Our TSNet significantly outperforms the other state-of-the-art methods. The lower the better. }
	\centering	
		\resizebox{1\linewidth}{!}{
			\setlength{\tabcolsep}{1.0em}%
			\begin{tabular}{c|ccccc|ccccc} 
				\toprule
				\multirow{3}{*}{Method}  &   \multicolumn{5}{c}{\textbf{PIE}}           & \multicolumn{5}{c}{\textbf{JAAD}}     \\
				\cline{2-11}
				~& \multicolumn{3}{c}{ADE}  &  C-ADE  &  C-FDE & \multicolumn{3}{c}{ADE}  &  C-ADE  &  C-FDE \\
				
				~ & 0.5s & 1.0s & 1.5s & 1.5s & ~ & 0.5s & 1.0s & 1.5s & 1.5s & ~ \\
				\midrule
				Linear~\cite{rasouli2019pie} & 123 & 477 & 1365 & 950 & 3983 & 233 & 857 & 2303 & 1565 & 6111 \\
				LSTM~\cite{rasouli2019pie} & 172 & 330 & 911 & 837 & 3352 & 289 & 569 & 1558 & 1473 & 5766 \\
				B-LSTM~\cite{bhattacharyya2018long} & 101 & 296 & 855 & 811 & 3259 & 159 & 539 & 1535 & 1447 & 5615 \\
				FOL-X~\cite{yao2019egocentric} & 47 & 183 & 584 & 546 & 2303 & 147 & 484 & 1374 & 1290 & 4924 \\
				$\text{PIE}_{traj}$~\cite{rasouli2019pie}& 58 & 200 & 636 & 596 & 2477 & 110 & 399 & 1280 & 1183 & 4780 \\
				BiTraP~\cite{yao2021bitrap} & 23 & 48 & 102 & 81 & 261 & \textbf{38} & 94 & 222 & 177 & 565 \\
				ABC+~\cite{halawa2022action} & 16 & 38 & 87 & 65 & 191 & 40 & 89 & 189 & 145 & 409 \\
				\hline
				TSNet & \textbf{15} & \textbf{34} & \textbf{73} & \textbf{51} & \textbf{133} & 41 & \textbf{84} & \textbf{166} & \textbf{121} & \textbf{325} \\
				\bottomrule
			\end{tabular}
		}
	
	\label{tab:1}
\end{table*}

\textbf{Final Trajectory Clustering.}
The multimodal trajectory prediction aims to predict $K$ possible trajectories to cover the ground truth.
Our proposed network can generate multi-modal trajectories with a CVAE module. However, when only limited samples are generated from the latent distribution, bias issues may arise because some samples may fall into low-density regions or too many samples may be gathered in high-density regions~\cite{xu2022socialvae}. Hence, we adopt the final trajectory clustering strategy to make the samples evenly distributed in each region of the latent distribution. 
Specifically, we first sample $C$~($C>>K$) latent variables $Z$ by the CVAE module and then generate $C$ corresponding goals $G$.
Subsequently, we predict $C$ trajectories $\hat{Y}$ conditioned by $C$ generated goals. Finally, we cluster the number of predicted trajectories from $C$ into $K$ as our final multimodal predictions.
Experiments show that the final trajectory clustering can improve performance, but it could cause an increment in the inference time.
We believe that an appropriate value of $C$ can be chosen to achieve a balance between the performance and the inference time.


\begin{table}[t]
	\begin{center}
		\resizebox{1\linewidth}{!}{
			\setlength{\tabcolsep}{0.6em}%
			\begin{tabular}{ccc|ccccc} 
				\hline
				\multirow{2}{*}{ST}&  \multirow{2}{*}{SC}& \multirow{2}{*}{FTC}&    \multicolumn{3}{c}{ADE}  &  C-ADE  &  C-FDE \\
				
				~ &&& 0.5s & 1.0s & 1.5s & 1.5s & ~ \\
				\hline
				\ding{53}&\ding{53} &\ding{53} & 18 & 45 & 111 & 85 & 285 \\
				\checkmark& \ding{53}&\ding{53}&17 & 42 & 98 & 77 & 245 \\
				\ding{53}&\checkmark& \ding{53}& 17 & 43 & 103 & 80 & 276  \\
				\checkmark&\checkmark&\ding{53}& 16 & 40 & 96 & 72 & 234 \\
				\checkmark&\checkmark&\checkmark& \textbf{15} & \textbf{34} & \textbf{73} & \textbf{51} & \textbf{133} \\
				\hline
			\end{tabular}
		}
	\end{center}
	\caption{Contribution of Each Component on the PIE dataset. The lower the better.}
	\label{tab:2}
\end{table}

\section{Experimental Analysis}
\label{Experimental Analysis}
We perform extensive experiments and compare experimental results with previous works on PIE~\cite{rasouli2019pie} and JAAD~\cite{8265243}. Moreover, we conduct comprehensive ablation studies to verify our main contributions.
\subsection{Datasets}
\textbf{PIE Dataset. }
The Pedestrian Intention Estimation~(PIE) dataset~\cite{rasouli2019pie} is a large-scale first-person view dataset that is captured from dash cameras annotated at 30Hz.
The PIE dataset contains 293,437 annotated frames and 1,842 pedestrians with annotated pedestrian character classes such as action, gesture, cross, look, age, gender, etc.
For example, the pedestrian character class ``action" has annotations for walking and standing, and the pedestrian character class ``age" has annotations for child, young, adult and senior.
We use five pedestrian character classes, \textit{i.e.}, action, gesture, look, gender and age, in our work.
We use the same training and testing splits as~\cite{rasouli2019pie, halawa2022action}.

\textbf{JAAD Dataset. }
The Joint Attention for Autonomous Driving~(JAAD) dataset~\cite{8265243} is a first-person view dataset that is captured from dash cameras annotated at 30Hz.
The JAAD dataset contains 82,032 annotated frames and 2,786 pedestrians, where 686 of them have pedestrian character annotations as the PIE dataset.
For pedestrians with no pedestrian character annotations, we manually annotate them as ``unknown".
We use five same pedestrian character classes as the PIE dataset in our work and the same training and testing splits as~\cite{rasouli2019pie, halawa2022action}.

\subsection{Experimental Setup}
\textbf{Implementation Details. }
The number of layers and the embedding dimension of self-attention in the sparse character representation stream are 1 and 64 respectively. The number of layers in GCN is $1$. The threshold $\xi$ is set to 0.5. The embedding dimension of encoders in the trajectory representation stream and the decoder in the decoder module are all set to 256. The hyper-parameter $C$ of final trajectory clustering is set to $100$. Our model is trained with batch size 128, learning rate 0.001, and an exponential LR scheduler~\cite{salzmann2020trajectron++}. 
Following previous methods~\cite{halawa2022action, yao2021bitrap, rasouli2019pie, 8265243}, we observed $0.5$ seconds and predict $0.5$, $1.0$ and $1.5$ seconds respectively.
We use the best-of-20 strategy for the multimodal prediction as previous methods~\cite{halawa2022action, yao2021bitrap, rasouli2019pie, 8265243}.
The entire framework is trained on GTX-3090 GPU.
All models are implemented with PyTorch.

\begin{table}[t]
	\begin{center}
		\resizebox{1\linewidth}{!}{
			\setlength{\tabcolsep}{0.6em}%
			\begin{tabular}{ccc|ccccc} 
				\hline
				\multirow{2}{*}{ST}&  \multirow{2}{*}{SC}& \multirow{2}{*}{FTC}&  \multicolumn{3}{c}{ADE}  &  C-ADE  &  C-FDE           \\
				
				~ &&& 0.5s & 1.0s & 1.5s & 1.5s & ~ \\
				\hline
				\ding{53}&\ding{53} &\ding{53} & 43 & 100 & 225 & 177 & 542 \\
				\checkmark& \ding{53}&\ding{53}&43 & 98 & 216 & 171 & 522 \\
				\ding{53}&\checkmark& \ding{53}& 43 & 99 & 217 & 170 & 518  \\
				\checkmark&\checkmark&\ding{53}& 43 & 96 & 202 & 155 & 469 \\
				\checkmark&\checkmark&\checkmark&  \textbf{41} & \textbf{84} & \textbf{166} & \textbf{121} & \textbf{325} \\
				\hline
			\end{tabular}
		}
	\end{center}
	\caption{Contribution of Each Component on the JAAD dataset. The lower the better.}
	\label{tab:3}
\end{table}

\begin{table*}[t]
	\begin{center}
		\resizebox{1\linewidth}{!}{
			\setlength{\tabcolsep}{1.0em}%
			\begin{tabular}{c|c|ccccc|ccccc} 
				\toprule
				\multirow{3}{*}{K}&\multirow{3}{*}{Method}&     \multicolumn{5}{c}{\textbf{PIE}}           & \multicolumn{5}{c}{\textbf{JAAD}}\\
				\cline{3-12}
				~&~&\multicolumn{3}{c}{ADE}  &  C-ADE  &  C-FDE &     \multicolumn{3}{c}{ADE}  &  C-ADE  &  C-FDE  \\
				&& 0.5s & 1.0s & 1.5s & 1.5s & ~ & 0.5s & 1.0s & 1.5s & 1.5s & ~\\
				\hline
				\multirow{2}{*}{15}&Bitrap\cite{yao2021bitrap} & 17 & 46 & 119 & 94 & 343 & 45 & 113 & 272 & 221 & 736\\
				&TSNet&\textbf{16} & \textbf{37} & \textbf{82} & \textbf{59} & \textbf{176} &\textbf{42} & \textbf{90} & \textbf{183} & \textbf{137} & \textbf{387}\\
				\hline
				\multirow{2}{*}{10}&Bitrap\cite{yao2021bitrap}& 19 & 58 & 164 & 137 & 541  & 49 & 140 & 367 & 313 & 1128\\
				&TSNet& \textbf{17} & \textbf{45} & \textbf{110} & \textbf{86} & \textbf{295} & \textbf{44} & \textbf{100} & \textbf{219} & \textbf{171} & \textbf{529}\\
				\hline
				\multirow{2}{*}{5}&Bitrap\cite{yao2021bitrap}& 27 & 111 & 378 & 345 & 1506 & 64 & 238 & 738 & 675 & 2710\\
				&TSNet& \textbf{22} & \textbf{83} & \textbf{259} & \textbf{232} & \textbf{958} & \textbf{51} & \textbf{144} & \textbf{379} & \textbf{329} & \textbf{1221}\\
				\bottomrule
			\end{tabular}
		}
	\end{center}
	\caption{Different best-of-K prediction on the PIE and JAAD dataset. All approaches input 15 frames and output 45 frames. The lower the better.}
	\label{tab:4}
\end{table*}
\begin{table}[t]
	\begin{center}
		\resizebox{1\linewidth}{!}{
			\setlength{\tabcolsep}{0.6em}%
			\begin{tabular}{cc|cc} 
				\hline
				\multirow{1}{*}{Dataset}&  \multirow{1}{*}{Character}  &  C-ADE~(1.5s)  &  C-FDE~(1.5s)            \\
				
				\hline
				\multirow{5}{*}{PIE} & Baseline & 85 & 285 \\
				~&Action& 79 & 258 \\
				~& Gesture&  80 & 267 \\
				~& Look&  \underline{86} & \underline{310}  \\
				~&Gender& 76 & \textbf{245} \\
				~&Age& \textbf{75} & 255 \\
				\hline
			\end{tabular}
		}
	\end{center}
	\caption{Impacts of different pedestrian character categories on the PIE dataset. Bold indicates the best performance. Underline indicates the worst performance. The lower the better.}
	\label{tab:5}
\end{table}

\begin{table}[t]
	\begin{center}
		\resizebox{1\linewidth}{!}{
			\setlength{\tabcolsep}{0.6em}%
			\begin{tabular}{cc|cc} 
				\hline
				\multirow{1}{*}{Dataset}&  \multirow{1}{*}{Character}  &  C-ADE~(1.5s)  &  C-FDE~(1.5s)            \\
				
				\hline
				\multirow{5}{*}{JAAD} & Baseline & 177 &  542 \\
				~&Action& \textbf{161} & \textbf{502} \\
				~& Gesture&  185 & 577 \\
				~& Look&  174 & 539  \\
				~&Gender& \underline{202} & \underline{638} \\
				~&Age& 168 & 509 \\
				%
				\hline
			\end{tabular}
		}
	\end{center}
	\caption{Impacts of different pedestrian character categories on the JAAD dataset. Bold indicates the best performance. Underline indicates the worst performance. The lower the better.}
	\label{tab:6}
\end{table}

\textbf{Evaluation Metrics. }
Following commonly accepted metrics~\cite{halawa2022action, yao2021bitrap}, we evaluate our method using:
(1) \textit{Bounding Box Average Displacement Error (ADE)}, which denotes the mean square error (MSE) distance of bounding box for the prediction and the ground truth;
(2) \textit{Bounding Box Center ADE (C-ADE)}, which denotes the MSE distance of bounding box center for the prediction and the ground truth;
(3) \textit{Bounding Box Final Displacement Error (FDE)}, which denotes the MSE distance between the destination bounding box for the prediction and the ground truth;
(4) \textit{Bounding Box Center FDE (C-FDE)}, which denotes the MSE distance between the destination bounding box center for the prediction and the ground truth.
The error of the bounding box is calculated using the upper-left and lower-right coordinates.

\subsection{Quantitative Analysis}
As shown in Table~\ref{tab:1}, we compare our method with seven first-person view trajectory prediction models, including Linear~\cite{rasouli2019pie}, LSTM~\cite{rasouli2019pie}, B-LSTM~\cite{bhattacharyya2018long}, FOL-X~\cite{yao2019egocentric}, $\text{PIE}_{traj}$~\cite{rasouli2019pie}, BiTraP~\cite{yao2021bitrap}, ABC+~\cite{halawa2022action}, on PIE~\cite{rasouli2019pie} and JAAD~\cite{8265243} datasets, where ABC+~\cite{halawa2022action} achieves the best performance among all state-of-the-art methods. 
The comparison with state-of-the-art methods indicates that our method significantly outperforms all other approaches on PIE and JAAD datasets. Compared with ABC+~\cite{halawa2022action}, \textit{i.e.}, the best method using pedestrian characters without sparsity, our method surpasses it by $21\%$~(C-ADE) and $30\%$~(C-FDE) on PIE and $16\%$~(C-ADE) and $20\%$~(C-FDE) on JAAD, which indicates the superiority of our method.

Moreover, the results indicate that our method has better performance on long-term prediction, which is proved by ADE results at different time steps in Table~\ref{tab:1}. Compared with ABC+~\cite{halawa2022action} on the PIE dataset, our method improves the performance by $6\%$, $10\%$ and $17\%$ on $0.5$s, $1.0$s and $1.5$s, respectively. Compared with ABC+~\cite{halawa2022action} on the JAAD dataset, our method improves the performance by $5\%$ and $12\%$ on $1.0s$ and $1.5$s, respectively. The underlying reason could be that our method removes temporal side-effect information due to long time distances from the future.

\begin{figure*}[t]
	\centering
	\includegraphics[width=0.8\textwidth]{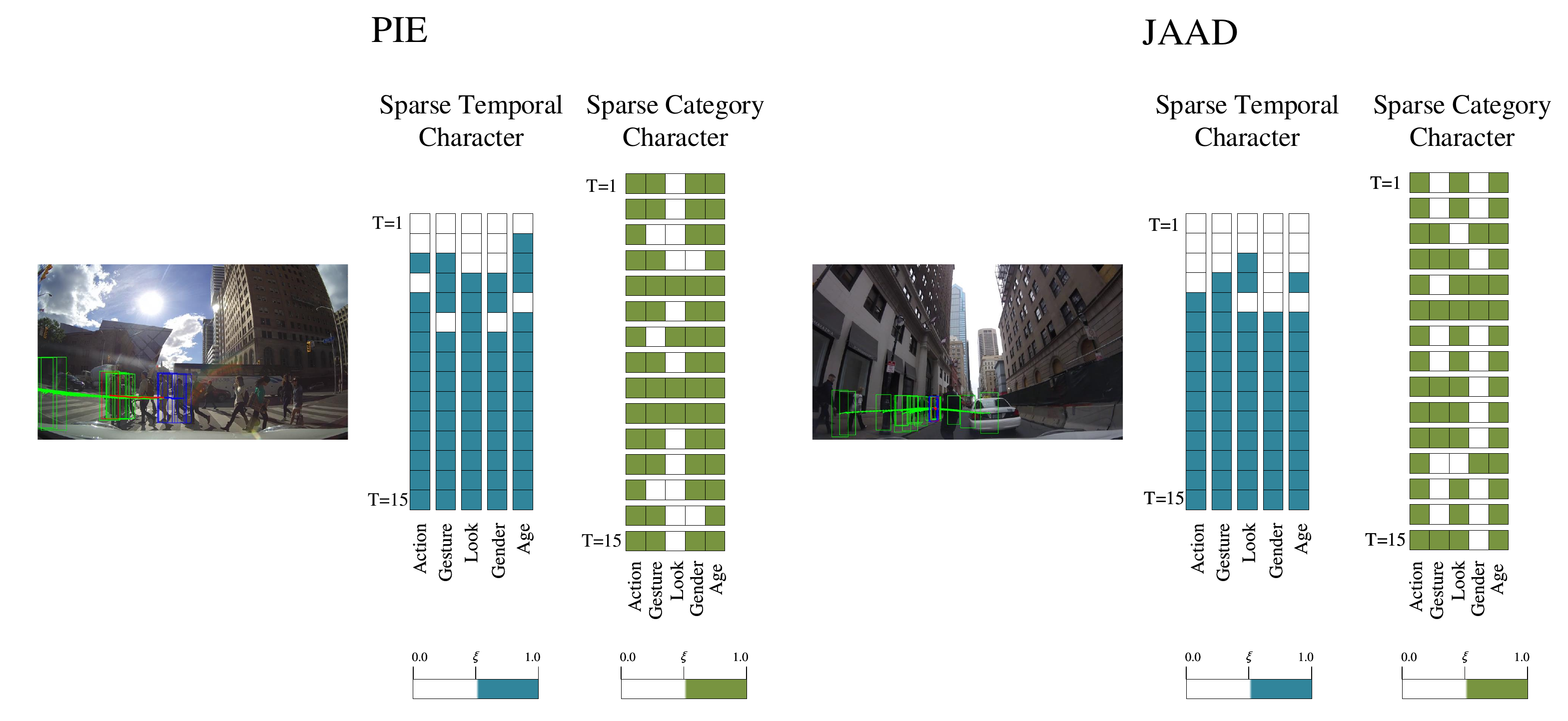}	
	\caption{  
		Examples of visualization from the PIE and JAAD datasets showing the sparse character of pedestrian.
		White blocks denotes removed characters. $\xi$ denotes the mask threshold.
	}
	\label{Mask}
\end{figure*}

\subsection{Ablation Study}
In this section, we conduct five ablation experiments.
Firstly, we remove each component of our proposed network to demonstrate their contributions.
Secondly, we conduct different best-of-K prediction to demonstrate the adaptability of our proposed method.
Thirdly, we replace multi-category pedestrian characters with diverse single-category characters to show varying representation abilities of different categories' pedestrian character information.
Fourthly, we change the mask threshold value $\xi$ to evaluate the performance of different degrees of sparsity.
Finally, we analyze the effectiveness of the final trajectory clustering.

\textbf{Contribution of Each Component. }
As illustrated in Table~\ref{tab:2} and Table~\ref{tab:3}, we evaluate the contributions of three components in our network, 
\textit{i.e.}, (1) ST denotes the sparse temporal character graph; (2) SC denotes the sparse category character graph; (3) FTC denotes the final trajectory clustering strategy.
The experiment results show that each component can lead to performance improvement.
Specifically, compared with the baseline, \textit{i.e.}, the method without all three components, adding the sparse temporal character graph can improve performance by $9\%$~(C-ADE) and $14\%$~(C-FDE) on the PIE dataset and $3\%$~(C-ADE) and $3\%$~(C-FDE) on the JAAD dataset;
adding the sparse category character graph can improve performance by $6\%$~(C-ADE) and $3\%$~(C-FDE) on the PIE dataset and $4\%$~(C-ADE) and $4\%$~(C-FDE) on the JAAD dataset;
adding all two sparse character graphs can improve performance by $15\%$~(C-ADE) and $18\%$~(C-FDE) on the PIE dataset and $12\%$~(C-ADE) and $13\%$~(C-FDE) on the JAAD dataset.
The results indicate the importance of both temporal character sparsity and category character sparsity for pedestrian trajectory prediction.

\begin{table}[t]
	\begin{center}
		\resizebox{1\linewidth}{!}{
			\setlength{\tabcolsep}{0.5em}%
			\begin{tabular}{c|c|ccccc} 
				\hline
				\multirow{2}{*}{Dataset}&\multirow{2}{*}{$\xi$}&     \multicolumn{3}{c}{ADE}  &  C-ADE  &  C-FDE \\
				
				~ & ~ & 0.5s & 1.0s & 1.5s & 1.5s & ~ \\
				\hline
				\multirow{4}{*}{PIE} & 0 & 17 & 38 & 81 & 59 & 162 \\
				~ &0.25 &17 & 36 & 74 & 51 & 138 \\
				~ &0.50&\textbf{15} & \textbf{34} & \textbf{73} & \textbf{51} & \textbf{133} \\
				~ &0.75& 16 & 37 & 81 & 58 & 152 \\
				\hline
			\end{tabular}
		}
	\end{center}
	\caption{Ablation study of threshold $\xi$ on the PIE dataset. Bold indicates the best performance. The lower the better.}
	\label{tab:9}
\end{table}
\begin{table}[t]
	\begin{center}
		\resizebox{1\linewidth}{!}{
			\setlength{\tabcolsep}{0.5em}%
			\begin{tabular}{c|c|ccccc} 
				\hline
				\multirow{2}{*}{Dataset}&\multirow{2}{*}{$\xi$}&     \multicolumn{3}{c}{ADE}  &  C-ADE  &  C-FDE \\
				
				~ & ~ & 0.5s & 1.0s & 1.5s & 1.5s & ~ \\
				\hline
				\multirow{4}{*}{JAAD} & 0 & 41 & 87 & 176 & 131 & 359 \\
				~ &0.25 &41 & 89 & 188 & 144 & 441 \\
				~ &0.50& \textbf{41} & \textbf{84} & \textbf{166} & \textbf{121} & \textbf{325}  \\
				~ &0.75& 42 & 92 & 189 & 143 & 396 \\
				%
				\hline
			\end{tabular}
		}
	\end{center}
	\caption{Ablation study of threshold $\xi$ on the JAAD dataset. Bold indicates the best performance. The lower the better.}
	\label{tab:10}
\end{table}

\begin{table}[t]
	\begin{center}
		\resizebox{1\linewidth}{!}{
			\setlength{\tabcolsep}{0.36em}%
			\begin{tabular}{cc|ccccc} 
				\hline
				\multirow{2}{*}{Dataset}&\multirow{2}{*}{\makecell[c]{Samples $C$}}	&     \multicolumn{3}{c}{ADE}  &  C-ADE  &  C-FDE \\
				~ &~& 0.5s & 1.0s & 1.5s & 1.5s & ~ \\
				\hline
				\multirow{5}{*}{PIE}&$20$ & 16 & 41 & 97 & 74 & 237 \\
				~ &	$40$&15 & 36 & 80 & 58 & 167 \\
				~ &$60$& 15 & 35 & 75 & 53 & 146  \\
				~ &$80$& 15 & 35 & 74 & 52 & 142 \\
				~ &$100$& 15 & 34 & 73 & 51 & 133 \\
				\hline
			\end{tabular}
		}
	\end{center}
	\caption{Ablation study of final trajectory clustering on the PIE dataset. The lower the better.}
	\label{tab:11}
\end{table}
\begin{table}[t]
	\begin{center}
		\resizebox{1\linewidth}{!}{
			\setlength{\tabcolsep}{0.36em}%
			\begin{tabular}{cc|ccccc} 
				\hline
				\multirow{2}{*}{Dataset} & \multirow{2}{*}{\makecell[c]{Samples $C$}}&     \multicolumn{3}{c}{ADE}  &  C-ADE  &  C-FDE \\
				~ && 0.5s & 1.0s & 1.5s & 1.5s & ~ \\
				\hline
				\multirow{5}{*}{JAAD}&$20$ & 43 & 96 & 205 & 157 & 469 \\
				~ &$40$& 42& 88 & 179 & 134 & 375 \\
				~ &$60$& 41 & 85 & 169 & 124 & 338 \\
				~ &$80$& 41 & 86 & 169 & 124 & 329 \\
				~ &$100$& 41 & 84 & 166 & 121 & 325 \\
				\hline
			\end{tabular}
		}
	\end{center}
	\caption{Ablation study of final trajectory clustering on the JAAD dataset. The lower the better.}
	\label{tab:12}
\end{table}

\textbf{Different Best-of-K Prediction}
Previous approaches commonly use Best-of-$K$ ($K=20$) as the quantified metric of multimodal trajectory prediction. To further validate the adaptability and effectiveness of our proposed TSNet, we conduct an experiment on various best-of-$K$ predictions with $K=5,10,15$ on PIE and JAAD datasets as shown in Table~\ref{tab:4}.
Due to ABC\cite{halawa2022action} have not released the code, we compare our method with the second-best performance approach Bitrap\cite{yao2021bitrap}.
Experiments results show that we outperform Bitrap\cite{yao2021bitrap} for different $K$, which indicates the adaptability and effectiveness of our proposed method.

\begin{figure*}[t]
	\centering
	\includegraphics[width=1.0\textwidth]{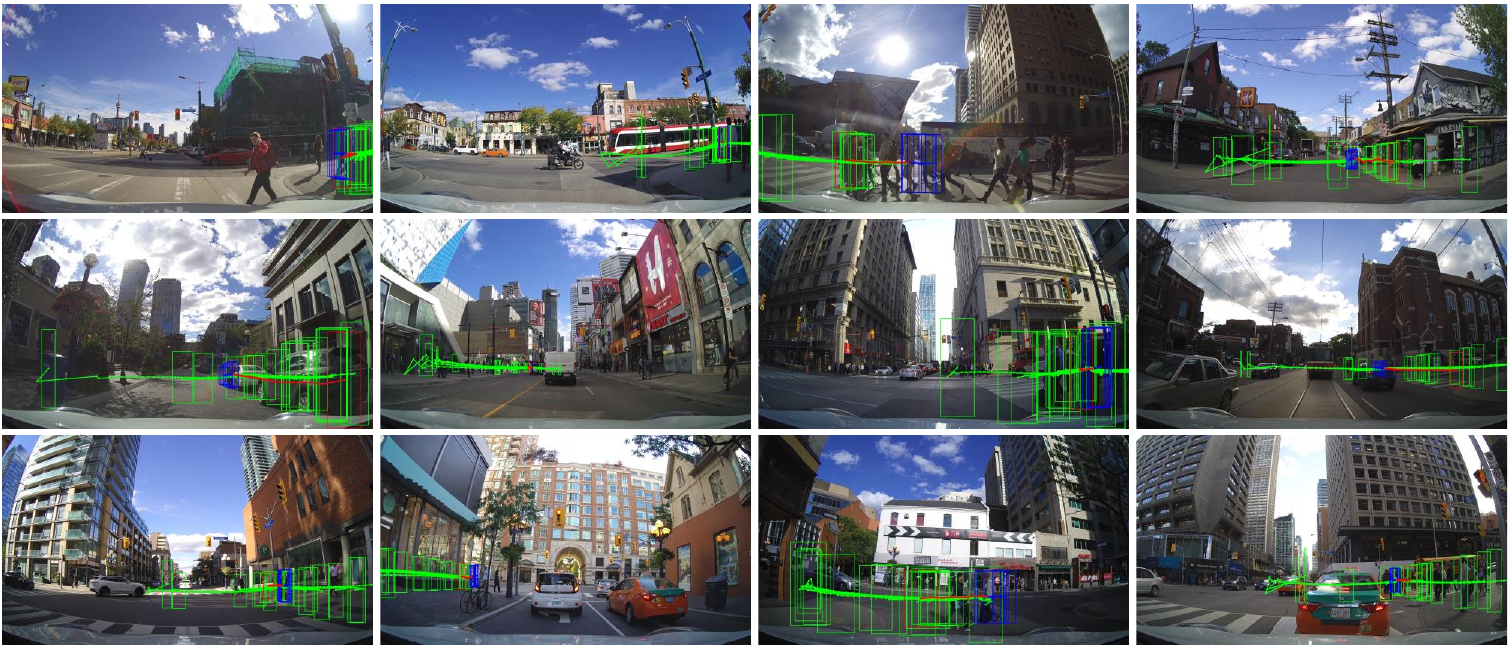}	
	\caption{  
		Examples of visualization from the PIE dataset showing the multimodality in the trajectory prediction space. 
		Blue bounding boxes are observed trajectories.
		Green bounding boxes are the multimodal predictions. 
		Red bounding boxes refer to the ground truth of future trajectories.
	}
	\label{PIE}
\end{figure*}
\begin{figure*}[t]
	\centering
	\includegraphics[width=1.0\textwidth]{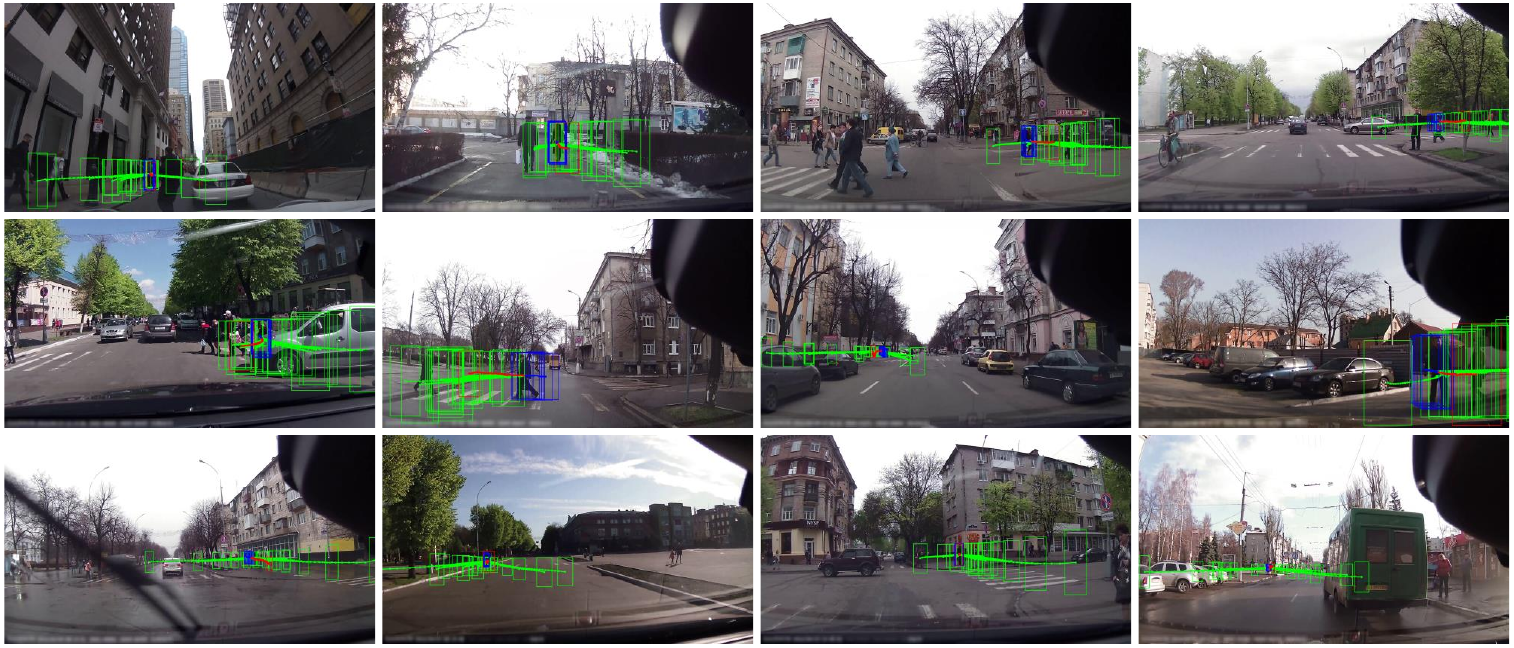}	
	\caption{  
		Examples of visualization from the JAAD dataset showing the multimodality in the trajectory prediction space. 
		Blue bounding boxes are observed trajectories.
		Green bounding boxes are the multimodal predictions. 
		Red bounding boxes refer to the ground truth of future trajectories.
	}
	\label{JAAD}
\end{figure*}

\textbf{Impacts of Different Pedestrian Character Information. }
As illustrated in Table~\ref{tab:5} and Table~\ref{tab:6}, we conduct experiments on using different single-category character information, including action, gesture, look, gender and age, to demonstrate that different categories of pedestrian characters have different relevance to future modes on PIE and JAAD datasets. The baseline denotes the model without using pedestrian character information.
Specifically, on the PIE dataset, the best performance is achieved on C-ADE and C-FDE when using ``age" and ``gender", respectively.
However, experiment results achieve the worst performance and even decrease compared with the baseline when using ``look".
On the JAAD dataset, the best performance is achieved on C-ADE and C-FDE when using ``action", and it causes a performance reduction compared with the baseline when using ``gender" and ``gesture".
The result indicates that different categories of pedestrian characters have different representation abilities due to their different relevance to future modes. Some kinds of pedestrian characters improve prediction accuracy to varying degrees, while some may cause adverse influence. 
Hence, it is essential to remove the adverse pedestrian characters.

\textbf{Analysis of Mask Threshold. }
As illustrated in Table~\ref{tab:9} and Table~\ref{tab:10}, we evaluate the mask threshold $\xi$ with four different values, including $0$, $0.25$, $0.50$ and $0.75$, on PIE and JAAD datasets.
The mask threshold $\xi$ means that we will eliminate the pedestrian characters when their scores are lower than the mask threshold value.
$\xi = 0$ means that we do not eliminate any pedestrian character information.
When $\xi = 0.5$, experiment results achieve the best performance on both PIE and JAAD datasets.
However, $\xi = 0$ achieves the worst result on the PIE dataset and $\xi = 0.25$ achieves the worst result on the JAAD datasets.
We believe that the choice of the mask threshold value depends on the difference of various datasets, including different pedestrian character categories, annotation errors, and the ego-motion.

\textbf{Analysis of Final Trajectory Clustering.}
As illustrated in Table~\ref{tab:11} and Table~\ref{tab:12}, we analyze to choose different numbers of samples $C$, which are clustered into $K~(K=20)$ multimodal prediction in PIE and JAAD datasets.
Experiment results show that the clustering post-process benefits the performance.
The performance increases when the number of samples $C$ increasing from $20$ to $100$.
Moreover, the performance improvement rate decreases as the value of $C$ increases.
When $C>100$, the performance is almost unchanged.
We choose $C=100$ in this work.

\subsection{Qualitative Analysis}
\textbf{Sparse Character Visualization. }
We show the sparse temporal characters and sparse category characters visualization on PIE and JAAD datasets in Figure~\ref{Mask}.
The two visualized trajectories are randomly selected from PIE and JAAD datasets, respectively.
Characters are removed for their mask values lower than mask threshold $\xi$. 
The sparse temporal characters shows that some pedestrian characters at time steps far from prediction are invalid.
The sparse temporal characters shows that some pedestrian characters in ``Look" categories are invalid in the PIE dataset and some pedestrian characters in ``Gender" and ``Age" categories are invalid in the JAAD dataset.
The sparse character visualization demonstrate that our method can remove negative pedestrian characters to improve the trajectory prediction.

\textbf{Trajectory Prediction Visualization. }
We show the visualization results on PIE and JAAD datasets in Figure~\ref{PIE} and Figure~\ref{JAAD}, respectively.
The visualization results are set to predict twenty possible trajectories following previous approaches~\cite{halawa2022action, yao2021bitrap}.
We observe that the multi-modal trajectories predicted by our method can sufficiently cover the ground truth in various urban and rural traffic scenarios.
Moreover, comparing the top two figures of the first column in Figure~\ref{PIE}, both figures have similar observation histories but different future trajectories, \textit{i.e.}, the ground truth of the pedestrian trajectory in the top figure is a sharp turn, while the ground truth of the pedestrian trajectory in the bottom figure is going straight. Our method can make correct predictions in this case, which indicates that our learned sparse pedestrian character features can help to distinguish different trajectory modes, which can not be distinguished only using trajectory coordinates. 

\section{Conclusion}
\label{Conclusion}
In this paper, we introduce a two-stream sparse-character-based network that leverages negative-removed pedestrian characters to enhance the representation of trajectory coordinates. Additionally, we propose a novel sparse character graph, which consists of sparse temporal and sparse category graphs, to model the different effects of various pedestrian characters and eliminate invalid information by adaptive masks in the temporal and category dimensions, respectively.
Through extensive experimental evaluations, we demonstrate that our method achieves significant performance improvements compared to previous state-of-the-art approaches. Furthermore, our ablation results reveal that different pedestrian characters exhibit varying representation abilities based on their relevance to future trajectory modes, highlighting the importance of sparsity in eliminating negative pedestrian characters. Moreover, our qualitative results illustrate the successful prediction of future trajectories in diverse urban and rural traffic scenarios.
The observed improvements in our method can be attributed to the enhanced representation of pedestrian characters achieved through the removal of negative characters.



%

%


\ifCLASSOPTIONcaptionsoff
  \newpage
\fi




\bibliographystyle{IEEEtran}
\bibliography{ref}
\end{document}